\documentclass[letterpaper]{article} 
\usepackage{aaai24}  
\usepackage{times}  
\usepackage{helvet}  
\usepackage{courier}  
\usepackage[hyphens]{url}  
\usepackage{graphicx} 
\usepackage{natbib}  
\usepackage{caption} 
\usepackage{algorithm}
\usepackage{algorithmic}
\usepackage{amsmath}

\usepackage{newfloat}
\usepackage{listings}
\DeclareCaptionStyle{ruled}{labelfont=normalfont,labelsep=colon,strut=off} 
\floatstyle{ruled}
\newfloat{listing}{tb}{lst}{}
\floatname{listing}{Listing}
\usepackage{amsfonts}
\usepackage{bm}
\usepackage{booktabs}
\usepackage{multirow}

\title{Towards Unified AI Drug Discovery with Multiple Knowledge Modalities}
\author{
    Yizhen Luo\equalcontrib\textsuperscript{\rm 12}, Xing Yi Liu\equalcontrib\textsuperscript{\rm 1}, Kai Yang\textsuperscript{\rm 1}, Kui Huang\textsuperscript{\rm 3}, Massimo Hong\textsuperscript{\rm 2}, Jiahuan Zhang\textsuperscript{\rm 1},\\Yushuai Wu\textsuperscript{\rm 1}, Zaiqing Nie\textsuperscript{\rm 1}
}
\affiliations{
    \textsuperscript{\rm 1}Institute for AI Industry Research (AIR), Tsinghua University, Beijing, China \\
    \textsuperscript{\rm 2}Department of Computer Science and Technology, Tsinghua University, Beijing, China \\
    \textsuperscript{\rm 3}School of Software and Microelectronics, Peking University, Beijing, China \\

    \{yz-luo22, hongcd21\}@mails.tsinghua.edu.cn \quad liuxingyi99@gmail.com \quad \{yangkai, zaiqing\}@air.tsinghua.edu.cn \quad kui@stu.pku.edu.cn \quad \{zhangjh0501, wuyushuai\}@mail.tsinghua.edu.cn
    
}

\usepackage{bibentry}

\begin{document}

\maketitle

\begin{abstract}
In recent years, AI models that mine intrinsic patterns from molecular structures and protein sequences have shown promise in accelerating drug discovery. However, these methods partly lag behind real-world pharmaceutical approaches of human experts that additionally grasp structured knowledge from knowledge bases and unstructured knowledge from biomedical literature. To bridge this gap, we propose KEDD, a unified, end-to-end, and multimodal deep learning framework that optimally incorporates both structured and unstructured knowledge for vast AI drug discovery tasks. The framework first extracts underlying characteristics from heterogeneous inputs, and then applies multimodal fusion for accurate prediction. To mitigate the problem of missing modalities, we leverage multi-head sparse attention and a modality masking mechanism to extract relevant information robustly. Benefiting from integrated knowledge, our framework achieves a deeper understanding of molecule entities, brings significant improvements over state-of-the-art methods on a wide range of tasks and benchmarks, and reveals its promising potential in assisting real-world drug discovery.
\end{abstract}

\section{Introduction}

Drug discovery aims to design molecules or compounds that respond to a certain disease and reduce their potential side effects on patients \cite{drews2000drug, lomenick2011identification, pushpakom2019drug}. The understanding of molecules, which entails either drugs or proteins, and their interactions builds the foundation of novel drug discovery processes \cite{paul2021artificial}. Such biomedical expertise usually resides within three different modalities: molecular structures, structured knowledge from knowledge graphs \cite{chaudhri2022knowledge}, and unstructured knowledge from biomedical documents \cite{saxena2022large}. These modalities complement each other, providing a holistic view to guide biomedical researchers.


While AI models that mine intrinsic patterns from molecular structures and protein sequences \cite{liu2021pre, wang2022molecular, zeng2022deep, rives2021biological} has achieved great success in assisting drug discovery, recent advances of multimodal models have shown the benefits of incorporating structured and unstructured knowledge in numerous downstream applications, including drug-target interaction prediction \cite{thafar2020dtigems, ye2021unified, yu2022hgdti}, drug-drug interaction prediction \cite{asada2018enhancing, zhang2017predicting, lin2020kgnn}, and protein-protein interaction prediction \cite{lv2021learning, zhang2021ontoprotein}. However, existing models are mostly restricted to a single task, and none of them attempt to take advantage of both structured and unstructured knowledge. This limits not only the application scope but also the capability of AI systems to holistically understand the intrinsic properties and functions of molecules. Besides, structured knowledge is occasionally unavailable for newly discovered molecules and proteins due to extensive cost of manual annotations, posing challenges of missing modality.

In this work, we propose KEDD, a unified end-to-end deep learning framework for \textbf{K}nowledge-\textbf{E}mpowered \textbf{D}rug \textbf{D}iscovery to solve the aforementioned problems. KEDD simultaneously harvests biomedical expertise from molecular structures, structured knowledge, and unstructured knowledge. KEDD could be flexibly applied to a wide range of AI drug discovery tasks. The framework first extracts unimodal features with independent encoders, and then performs modality fusion for accurate predictions. To alleviate the missing structured knowledge problem, KEDD leverages multi-head sparse attention to extract the most relevant information from knowledge bases, and improves the training of sparse attention with a modality masking mechanism. 


Comprehensive experiments on numerous AI drug discovery benchmarks demonstrate KEDD's capability of jointly comprehending and reasoning over different modalities. KEDD outperforms state-of-the-art models by an average of 5.2\% on drug-target interaction prediction, 3.4\% on drug property prediction, 1.2\% on drug-drug interaction prediction, and 4.1\% on protein-protein interaction prediction. Additionally, our results shed light on KEDD’s joint comprehension of different modalities and its potential in assisting real-world drug discovery.

Our main contributions are summarized as follows:
\begin{itemize}
    \item We present KEDD, a unified, end-to-end framework incorporating a wealth of modalities, namely molecular, structured knowledge, and unstructured knowledge.
    \item We propose multi-head sparse attention and modality masking to alleviate the missing modality problem for structured knowledge.
    \item We demonstrate the state-of-the-art performance of KEDD in wide-ranging AI drug discovery tasks.
\end{itemize}

\section{Related Works}



\subsubsection{Knowledge-empowered deep learning in AI drug discovery.} The exposive amount of structured and unstructured knowledge have sparked a wide range of knowledge-empowered deep learning approaches. have attempted to incorporate . In drug-target interaction prediction (DTI), DTIGems+  \cite{thafar2020dtigems} leverages node2vec \cite{grover2016node2vec} embeddings and a drug–target path scorer to predict the interaction. KGE\_NFM \cite{ye2021unified} proposes to mitigate the cold-start problem by combining knowledge graph embeddings and molecular structure features. Differently, HGDTI \cite{yu2022hgdti} leverages a heterogeneous graph neural network for DTI classification. In DDI, structural characteristics are assisted by knowledge graphs \cite{zhang2017predicting, karim2019drug, lin2020kgnn, ren2022biomedical} or textual descriptions \cite{asada2018enhancing} in isolation to better identify the relationships between drugs. In protein-protein interaction prediction, the effectiveness of mining knowledge graphs is also validated \cite{lv2021learning, zhang2021ontoprotein}. While existing model have achieved promising results, none of them attempt to harvest the advantages of both structured and unstructured knowledge.   

\subsubsection{Missing modality in multimodal learning.} Missing modality is a common problem in real world scenarios, where data from one or more modalities is incomplete \cite{ma2021smil}. To solve this problem, numerous approaches have been proposed, including late fusion \cite{steyaert2023multimodal}, missing modality reconstruction \cite{zhou2019latent, ma2021smil}, specialized fusion architectures \cite{ma2022multimodal}, and prompting \cite{lee2023multimodal}. In AI drug discovery, drugs and proteins may lack structured knowledge within knowledge bases, raising the missing modality problem. KEDD serves as the first attempt to address this problem by reconstructing the missing modality with sparse attention.


\section{Method}
In this section, we start with a brief introduction of preliminaries and denotations. Then, we describe the overall architecture of KEDD. Finally, we introduce the sparse attention module and modality masking technique in detail.

\subsection{Preliminaries}

KEDD focuses on two types of molecules involved in drug discovery: drugs and proteins. Each component further consists of information from three modalities, namely molecular structure, structured knowledge, and unstructured knowledge. Formally:
\begin{equation}
\begin{aligned}
&D=(D_\text{S}, D_\text{SK}, D_\text{UK})\in \mathcal{D},\\
&P=(P_\text{S}, P_\text{SK}, P_\text{UK})\in \mathcal{P},
\end{aligned}
\end{equation}
where $D$ refers to a drug, $P$ refers to a protein, and $\mathcal{D}, \mathcal{P}$ refers to the drug and protein spaces. The drug structure $D_\text{S}$ is profiled as a 2D molecular graph $(\mathcal{V}, \mathcal{E})$, where $\mathcal{V}$ denotes atoms, and $\mathcal{E}$ denotes molecular bonds. The protein structure $P_\text{S}$ is profiled as an amino acid sequence $[p_1, p_2, \ldots, p_m]$. The structured knowledge $D_\text{SK}$ and $P_\text{SK}$ corresponds to an entity within a knowledge base. The unstructured knowledge $D_\text{UK}$ and $P_\text{UK}$ is encapsulated in a text sequence $[t_1, t_2, \cdots, t_L]$ of length $L$.

AI drug discovery tasks that mine properties and interactions between drugs and proteins could be formulated as learning mapping functions from the drug, protein, or joint spaces to binary values. Formally:
\begin{itemize}
    \item \textbf{Drug-target interaction prediction (DTI)} predicts the binding effects between . This sheds light on the ability of chemical compounds in drugs to affect desired targets in the human body. The task is formulated as learning $\mathcal{F}_\text{DTI}: \mathcal{D} \times \mathcal{P} \to \{0, 1\}$.
    \item \textbf{Drug property prediction (DP)} predicts the existence of certain molecular properties such as soluablity and toxicity, which plays a significant role in developing safe drugs. The task is formulated as learning $\mathcal{F}_\text{DP}: \mathcal{D} \to \{0, 1\}$.
    \item \textbf{Drug-drug interaction (DDI)}. DDI predicts the connection between two drugs, which is beneficial in designing combinational treatment of multiple drugs. The task is formulated as learning $\mathcal{F}_\text{DDI}: \mathcal{D} \times \mathcal{D} \to \{0, 1\}$.
    \item \textbf{Protein-protein interaction prediction (PPI)} predicting different types of interaction relationships between proteins, which is beneficial for identifying the functions and drug abilities of molecules \cite{jones1996principles}. The task is formulated as learning $\mathcal{F}_\text{PPI}: \mathcal{P} \times \mathcal{P} \to \{0, 1\}^{n}$, where $n$ is the number of relation types.
\end{itemize}

For DTI, DDI, and PPI, the binary output indicates if a specific type of interaction exists between the inputs. For DP, the binary output indicates if the molecule holds a specific property. Due their similar formulations, we endeavor to build a unified end-to-end deep learning framework to solve these tasks with minimal modifications.

\begin{figure*}[t]
\centering
\includegraphics[width=0.98\textwidth]{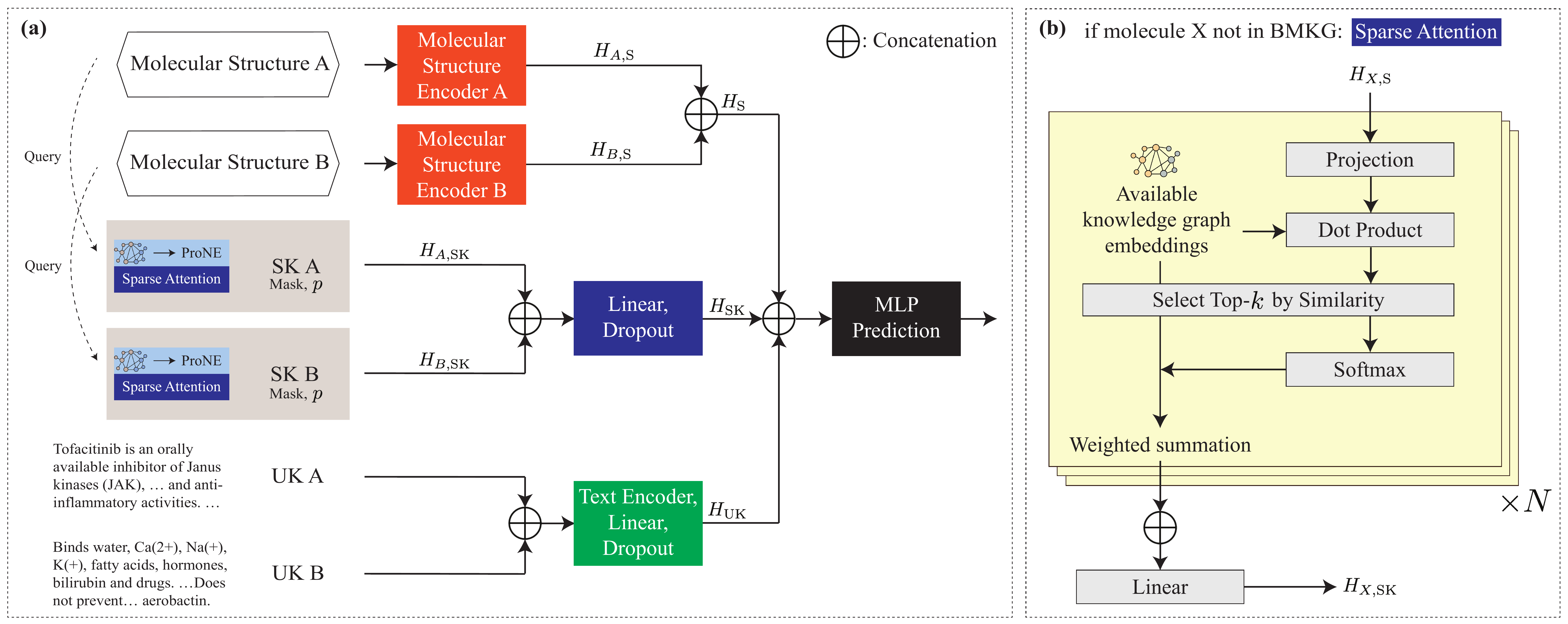} 
\caption{(a) The KEDD framework. GraphMVP and MCNN can both serve as molecular structure encoders A and/or B, depending on the task. The ``B" branches may also remain unused in the case of DP prediction. SK: structured knowledge; UK: unstructured knowledge. (b) Sparse attention pipeline for obtaining structured knowledge embeddings if a certain molecule is not found in BMKG.}
\label{fig:framework}
\end{figure*}


\subsection{KEDD Architecture}
Figure \ref{fig:framework}(a) illustrates the overall KEDD framework. Due to the heterogeneity between different modalities, we incorporate independent encoders to harvest biomedical expertise from each modality. Specifically: 

\begin{itemize}
    \item To encode a drug's molecular graph $D_\text{S}$, we use GraphMVP \cite{liu2021pre}, a 5-layer GIN \cite{xu2018powerful} pre-trained on both 2D molecular graphs and 3D molecular genomics. To encode protein structure $P_S$, we use multi-scale CNN (MCNN) \cite{yang2022mgraphdta}, a network with three distinct numbers of convolutional layers in each branch . Notably, the parameters of two molecular structure encoders are shared in DDI and PPI tasks. The molecular structure features $H_{A,S}, H_{B,S}$, processed either by GraphMVP or MCNN, are concatenated to formulate the overall structure feature $H_S$.
    
    \item We leverage ProNE \cite{zhang2019prone}, a fast and efficient network embedding algorithm, to harvest structured knowledge within knowledge graphs by incorporating relational and topological information. The embedding vectors for two molecules $H_{A,\text{SK}}, H_{B,\text{SK}}$ are concatenated and fed into a linear layer with dropout to formulate the structured knowledge feature $H_{SK}$. 
    
    \item We adopt PubMedBERT \cite{gu2021domain}, a language model pre-trained on biomedical corpus, to extract unstructured knowledge from noisy text descriptions. It is composed of 12 Transformer layers, and transforms a token sequence into contextualized embeddings. We adopt the outputs of the \texttt{[CLS]} token $H_i$, and feed it into a fully-connected layer with dropout to obtain unstructured knowledge feature $H_\text{UK}$. Notably, the textual descriptions of two molecules are concatenated with a \texttt{[SEP]} token before feeding them into PubMedBERT. Such a design enables the language model to better capture the cooccurrence of key information, thus supporting downstream relation prediction. 
\end{itemize}

Finally, the features from three modalities are concatenated, and passed into a multi-layer perceptron to generate prediction results. In the case of DP prediction, the branch for the second molecule simply produces empty vectors for each modality. We defer readers to the supplementary materials for detailed architecture of KEDD for each task.


\subsection{Mitigating Missing Modality with Sparse Attention and Modality Masking}


Ideally, each molecule is compiled with corresponding structured and unstructured knowledge to facilitate multimodal comprehension. However, in real-world drug discovery, a large portion of molecules, especially those that are newly discovered, could not be linked to knowledge bases due to extensive cost of manual annotations, posing challenges of missing modality for structured knowledge. 

To mitigate this problem, we leverage sparse attention \cite{zhao2019sparse} to compose the missing structured knowledge by querying the most relevant entities within the large-scale knowledge graph based on molecular structure. As illustrated in Figure 1(b), we project the molecular structure features to the feature space of structured knowledge. We use the projection results $\tilde{H}_{X,S}$ as queries, and the knowledge graph embedding matrix $E$ as keys and values. The sparse attention matrix $A$ is calculated by selecting top-$k$ relevant entities based on original attention scores:
\begin{equation}
\begin{aligned}
    &Q=W_Q\tilde{H}_{X,S}, K=W_KE, A=\frac{QK^T}{\sqrt{d}}\\
    &\tilde{A}=\mathrm{softmax}(\mathrm{Top}(P, k)),
\end{aligned}
\end{equation}
where $W_Q, W_K$ are trainable parameters, $\mathrm{Top}(P, k)$ selects $k$ largest elements within each row of P, and withdraws the remaining elements by assigning a similarity score of $-\infty$.  

Finally, the missing modality of structured knowledge is computed as follows:
\begin{equation}
    V=W_VE, H_{X,SK}=\tilde{A}V,
\end{equation}
where $W_V$ is defined as an identity matrix to ensure that $H_{X,SK}$ resides within the feature space of original knowledge embeddings.

On occasions where the missing modality problem is not too severe, the number of samples could be insufficient for the sparse attention module to elicit informative structured knowledge from the knowledge graph. To address this issue, we propose a modality mask strategy on structured knowledge inputs. With a probability of $p$, the available structured knowledge $H_{X,SK}$ for a molecule is masked, and the sparse attention is activated. The masked sample is trained on the original task-specific objective instead of reconstruction objectives to achieve a deeper understanding of the relationships between unstructured knowledge and drug discovery tasks. This strategy expands supervision signals for sparse attention, and improves the robustness of our framework since the sparse attention outputs could be viewed as a form of data augmentation for structured knowledge.



\section{Experiments and Results}

\subsection{Data preparation}
Since the majority of existing datasets for AI drug discovery only provide structural information for drugs and proteins, we supplement them with multimodal structured and unstructured knowledge extracted from public repositories \cite{boeckmann2003swiss, wishart2018drugbank, kanehisa2007kegg, zheng2021pharmkg, consortium2015uniprot}. We build BMKG, a dataset containing molecular structure, interacting relationships, and textual descriptions for 6,917 drugs and 19,992 proteins. In total, BMKG contains 2,223,850 drug-drug links, 47,530 drug-protein links and 633,696 protein-protein links. We obtain inputs for structured and unstructured knowledge by comparing the structural information of drugs and proteins for each dataset.

KEDD is applied on 4 popular downstream tasks with 9 benchmark datasets summarized in Table \ref{tab:datasets}.
\begin{table}[h]
  \centering
  \fontsize{9}{11}\selectfont
  \begin{tabular}{ c | c c c c }
    \toprule
    \textbf{Task} & \textbf{Dataset} & \textbf{\# Drugs} & \textbf{\# Proteins} & \textbf{\# Samples} \\
    \midrule
    \multirow{2}{*}{DTI} & BMKG-DTI & 2803/2803 & 2810/2810 & 47391 \\
    & Yamanishi08 & 488/791 & 944/989 & 10254 \\
    \midrule
    \multirow{4}{*}{DP} & BBBP & 841/2039 & - & 2039 \\
    & ClinTox & 556/1478 & - & 1478 \\
    & Tox21 & 2191/7831 & - & 7831 \\
    & SIDER & 677/1427 & - & 1427 \\
    \midrule
    DDI & Luo & 657/721 & - & 494551 \\
    \midrule
    \multirow{2}{*}{PPI} & SHS27k & - & 1632/1690 & 10928 \\
    & SHS148k & - & 4943/5189 & 63065 \\
    \bottomrule
  \end{tabular}
  \caption{A brief summary of benchmark datasets. The total number of molecules in the dataset is to the right of /, and the number of molecules linked to BMKG is to the left of /.}
  \label{tab:datasets}
\end{table}

\begin{itemize}
\item For DTI, we adopt two binary classification datasets, Yamanishi08 \cite{yamanishi2008prediction} and BMKG-DTI. The latter is extracted from BMKG, thus free from the missing modality problem. More details of this dataset are available in supplementary materials. We perform 5-fold cross validation for the warm, cold drug, and cold protein start settings, and 9-fold cross validation for the cold cluster start setting. Under the warm start setting, drug-protein pairs are randomly partitioned. Under the cold drug, cold protein, and cold cluster start settings, drugs, proteins, and both in the test set, respectively, are unseen during training. The cold start settings are more similar to real-world drug discovery, where researchers endeavour to figure out the binding effects between novel drugs and targets.

\item For DP, we select 4 representative binary classification datasets from MoleculeNet \cite{wu2018moleculenet}, a widely-adopted benchmark for molecular machine learning. The drug properties involves blood-brain barrier penetration, FDA approval status, toxicity, and side effects to multiple organs. Scaffold split \cite{wu2018moleculenet} with a train-validation-test ratio of 8:1:1 is applied, and AUROC is reported.

\item For DDI, we adopt Luo's dataset \cite{luo2017network}. We randomly split the binary classification dataset with a train-validation-test ratio of 8:1:1, and report AUROC and AUPR.

\item For PPI, we leverage the revised version of multi-label classification datasets SHS27k and SHS148k \cite{chen2019multifaceted}. We follow the BFS and DFS strategy in GNN-PPI \cite{lv2021learning} to split the dataset. We adopt Micro F1 score as the evaluation metric.
\end{itemize}

\subsection{Implementation Details}

Our sparse attention module composes 4 attention heads, and we set $k=16$ across our experiments. The modality masking probability $p$ is set with $0.05$ across most models. To avoid information leakage, we remove connections between drugs and proteins in the test set of DDI, DTI and PPI datasets from BMKG. Each KEDD model was trained on a single A100 40GB GPU using PyTorch, with a maximum training cost of 1 day. Each experiment is performed 3 times with different seeds. For more details of our pre-processing procedure and hyperparameters, please refer to supplementary materials.

\subsection{Performance Evaluation on Downstream Tasks}

\subsubsection{DTI.} 
\begin{figure}[h]
\centering
\includegraphics[width=0.98\columnwidth]{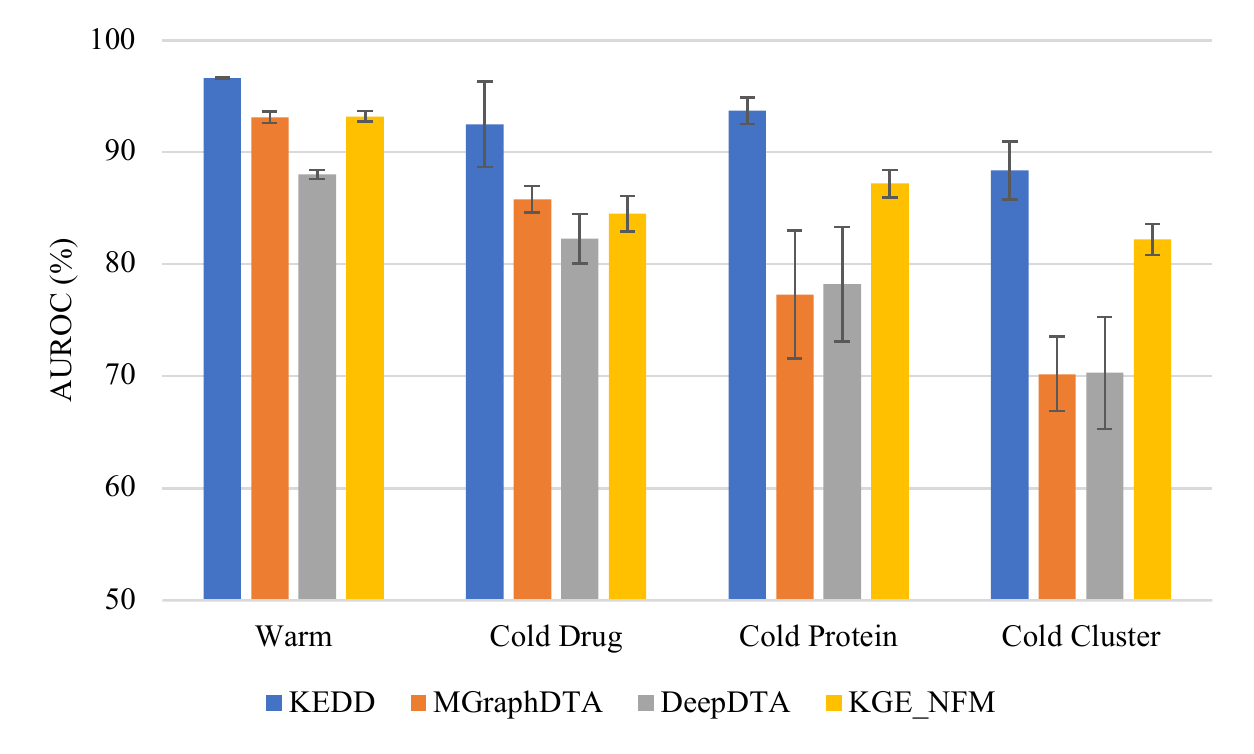}
\caption{AUROC on the BMKG-DTI dataset.} 
\label{fig:auroc-bmkgdti}
\end{figure}

We compare KEDD against state of the art methods including DeepDTA \cite{ozturk2018deepdta}, GraphDTA \cite{nguyen2021graphdta}, MGraphDTA \cite{yang2022mgraphdta}, SMT-DTA \cite{pei2022smt} and KGE\_NFM \cite{ye2021unified}. 
The AUROC results are shown in Figure \ref{fig:auroc-bmkgdti} and Figure \ref{fig:auroc-yamanishi08}. The complete experiment results are displayed in supplementary materials. 

\begin{figure}[h]
\centering
\includegraphics[width=0.98\columnwidth]{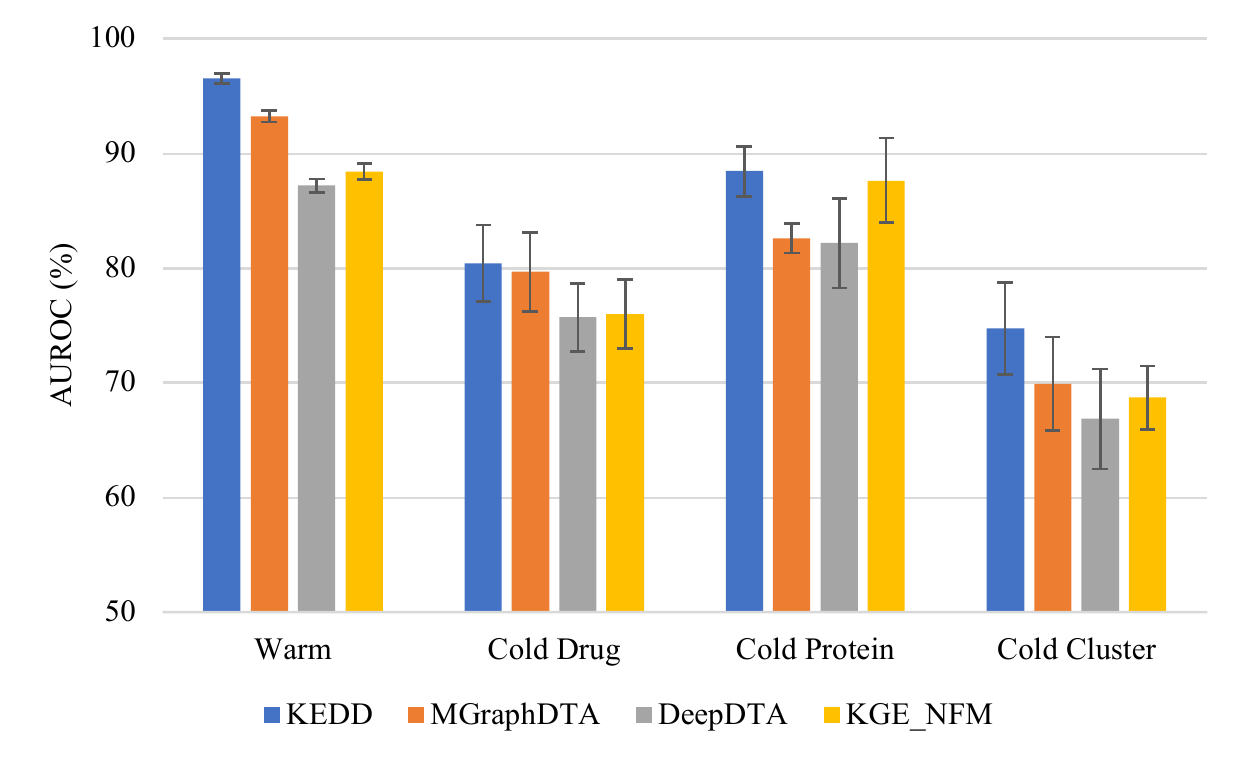}
\caption{AUROC on the Yamanishi08 dataset.} 
\label{fig:auroc-yamanishi08}
\end{figure}

\begin{table*}[t]
  \fontsize{9}{11}\selectfont
  \centering
  \begin{tabular}{ c | c c c c c }
    \toprule
    \textbf{Model} & \textbf{BBBP} & \textbf{ClinTox} & \textbf{SIDER} & \textbf{Tox21} & \textbf{Average} \\
    \midrule
    MolCLR & $71.1_{\pm 1.4}$ & $61.1_{\pm 3.6}$ & $57.7_{\pm 2.0}$ & $74.0_{\pm 1.0}$ & $65.9$ \\
    KV-PLM & $66.9_{\pm 1.1}$ & $84.3_{\pm 1.5}$ & $55.3_{\pm 0.9}$ & $64.7_{\pm 1.8}$ & $67.8$ \\
    MoMu & $70.5_{\pm 2.0}$ & $79.9_{\pm 4.1}$ & $60.5_{\pm 0.9}$ & $75.6_{\pm 0.3}$ & $71.6$ \\
    MoCL & $71.4_{\pm 1.1}$ & $81.4_{\pm 1.0}$ & $61.9_{\pm 0.4}$ & $72.5_{\pm 1.0}$ & $71.8$ \\
    GraphMVP & $72.4_{\pm 1.6}$ & $79.1_{\pm 2.8}$ & $63.9_{\pm 1.2}$ & $75.9_{\pm 0.5}$ & $72.8$ \\
    \midrule
    KEDD (w/o SK) & $71.7_{\pm 1.0}$ & $86.2_{\pm 2.9}$ & $61.9_{\pm 0.8}$ & $74.9_{\pm 0.5}$ & $73.7$ \\
    KEDD (w/o UK) & $71.2_{\pm 1.2}$ & $72.5_{\pm 6.4}$ & $63.9_{\pm 0.6}$ & $75.8_{\pm 0.3}$ & $70.8$ \\
    KEDD (w/o SA) & $71.3_{\pm 1.1}$ & $87.2_{\pm 1.3}$ & $62.8_{\pm 1.5}$ & $75.1_{\pm 1.0}$ & $74.1$ \\
    KEDD & \textbf{73.6}$_{\pm \textbf{1.1}}$ & \textbf{88.4}$_{\pm \textbf{0.7}}$ & \textbf{66.0}$_{\pm \textbf{1.4}}$ & \textbf{76.8}$_{\pm \textbf{0.4}}$ & \textbf{76.2} \\
    \bottomrule
  \end{tabular}
  \caption{Mean and standard deviation of AUROC (\%) on DP using four MoleculeNet datasets. w/o SK: without structured knowledge; w/o UK: without unstructured knowledge; w/o SA: without sparse attention.}
  \label{tab:dp}
\end{table*}





From the figures we observe that KEDD outperforms state-of-the-art models on both datasets. Compared to MGraphDTA, KEDD achieves a notable gain of 3.4\% and 3.5\% in AUROC under the warm start setting (paired $t$-test, $p$-value $<1.3 \times 10^{-6}$) on Yamanishi08 and BMKG-DTI. On cold-start scenarios that are more challenging, KEDD consistantly achieves superior performance except for the cold protein setting on Yamanishi08, where it shows minor statistical difference with KGE\_NFM (paired $t-$test, $p$-value $>0.05$). Notably, on BMKG-DTI where the missing modality problem does not exist, KEDD exhibits profound improvements over baselines with an average performance gain of 8.1\%, 7.5\%, 5.2\% on cold-drug, cold-protein and cold-cluster scenarios, respectively (paired $t-$tests, all $p$-values $< 2.9 \times 10^{-3}$). It even achives competitive results with that of warm start settings. These results demonstrate the benefits of incorporating structured and unstructured knowledge, especially for molecules that are out of the generalization scope of structure-based models.


\subsubsection{DP.} Comparisons between KEDD and MolCLR \cite{wang2022molecular}, KV-PLM \cite{zeng2022deep}, MoMu \cite{su2022molecular}, MoCL \cite{sun2021mocl}, and GraphMVP \cite{liu2021pre} are presented in Table \ref{tab:dp}. KEDD achieves state-of-the-art performance across all benchmarks, yielding an average improvement of 3.4\% in AUROC (paired $t-$test, $p$-value $<6.0 \times 10^{-2}$) by jointly reasoning over molecular structures, structured knowledge, and unstructured knowledge. 


\subsubsection{DDI.} For this task, we adopt baselines including DeepDTnet \cite{zeng2020target}, KGE\_NFM \cite{ye2021unified}, DTINet \cite{luo2017network}, DDIMDL \cite{deng2020multimodal}, DeepR2cov \cite{wang2021deepr2cov}, and MSSL2drug \cite{wang2023multitask}. As shown in Table \ref{tab:ddi}, KEDD achieves state-of-the-art results on the Luo dataset in both AUROC and AUPR. It also demonstrates robustness by achieving the least standard deviation between different runs. 
\begin{table}[h]
  \fontsize{9}{11}\selectfont
  \centering
  \begin{tabular}{ c | c c }
    \toprule
    \textbf{Model} & \textbf{AUROC (\%)} & \textbf{AUPR (\%)} \\
    \midrule
    DeepDTnet$^\dag$ & $92.3_{\pm 0.8}$ & $92.1_{\pm 1.0}$ \\
    KGE\_NFM$^\dag$ & $91.6_{\pm 0.8}$ & $90.7_{\pm 1.0}$ \\
    DTINet$^\dag$ & $92.9_{\pm 0.6}$ & $92.7_{\pm 0.9}$ \\
    DDIMDL$^\dag$ & $91.3_{\pm 0.9}$ & $90.5_{\pm 1.4}$ \\
    DeepR2cov$^\dag$ & $93.1_{\pm 0.9}$ & $91.2_{\pm 1.2}$ \\
    MSSL2drug$^\dag$ & $95.1_{\pm 0.4}$ & \textbf{94.4}$_{\pm \textbf{1.1}}$ \\
    \midrule
    KEDD (w/o SK) & $96.3_{\pm 0.1}$ & $91.7_{\pm 0.2}$ \\
    KEDD (w/o UK) & $97.1_{\pm 0.1}$ & $92.9_{\pm 0.2}$ \\
    KEDD (w/o SA) & $97.4_{\pm 0.1}$ & $94.1_{\pm 0.2}$ \\
    KEDD & \textbf{97.5}$_{\pm \textbf{0.1}}$ & \textbf{94.4}$_{\pm \textbf{0.2}}$ \\
    \bottomrule
  \end{tabular}
  \caption{Mean and standard deviation of AUROC and AUPR on DDI on Luo's dataset. $^\dag$: these results are taken from MSSL2drug \cite{wang2023multitask}. w/o SK: structured knowledge; w/o UK: unstructured knowledge; w/o SA: sparse attention.}
  \label{tab:ddi}
\end{table}



\subsubsection{PPI.} 

\begin{table}[h]
  \fontsize{9}{11}\selectfont
  \setlength{\tabcolsep}{3pt}
  \centering
  \begin{tabular}{ c | c c | c c}
    \toprule
    \multirow{2}{*}{\textbf{Model}} & \multicolumn{2}{c |}{\textbf{SHS27k}} & \multicolumn{2}{c}{\textbf{SHS148k}} \\
    & {DFS} & {BFS} & {DFS} & {BFS} \\
    \midrule
    PIPR & $53.0_{\pm 2.0}$ & $47.1_{\pm 2.4}$ & $56.5_{\pm 1.2}$ & $48.3_{\pm 0.7}$ \\
    GNN-PPI & $55.1_{\pm 1.1}$ & $52.4_{\pm 2.1}$ & $59.3_{\pm 0.9}$ & $44.8_{\pm 3.1}$ \\
    OntoProtein & $56.8_{\pm 0.4}$ & $61.2_{\pm 1.6}$ & $60.8_{\pm 0.8}$ & $48.0_{\pm 1.2}$ \\
    ESM-1b & $61.1_{\pm 1.0}$ & \textbf{62.9}$_{\pm \textbf{1.2}}$ & $63.2_{\pm 0.8}$ & $55.2_{\pm 0.5}$  \\
    \midrule
    KEDD (w/o SK) & $60.4_{\pm 1.5}$ & $55.6_{\pm 0.6}$ & $66.8_{\pm 1.2}$ & $55.0_{\pm 1.2}$ \\
    KEDD (w/o UK) & $62.8_{\pm 2.0}$ & $61.3_{\pm 1.0}$ & $68.2_{\pm 0.9}$ & $55.3_{\pm 0.8}$ \\
    KEDD (w/o SA) & $63.4_{\pm 1.3}$ & $62.3_{\pm 1.2}$ & $68.9_{\pm 0.8}$ & $57.2_{\pm 0.5}$ \\
    KEDD & \textbf{63.8}$_{\pm 1.5}$ & $62.7_{\pm 1.5}$ & \textbf{69.4}$_{\pm \textbf{1.0}}$ & \textbf{57.3}$_{\pm \textbf{1.1}}$ \\
    \bottomrule
  \end{tabular}
  \caption{Mean and standard deviation of F1 score (\%) on PPI using SHS148k dataset. w/o SK: without structured knowledge; w/o UK: without unstructured knowledge; w/o SA: without sparse attention.}
  \label{tab:ppi}
\end{table}

In Table \ref{tab:ppi}, we show the results of KEDD on the SHS148k dataset, compared against PIPR \cite{chen2019multifaceted}, GNN-PPI \cite{lv2021learning}, OntoProtein \cite{zhang2021ontoprotein}, and ESM-1b \cite{rives2021biological}. On SHS27k, KEDD outperforms baselines under the DFS setting (paired $t-$ test, $p$-value $<3.3 \times 10^{-2}$. Under the BFS setting, KEDD shows little statistical difference with ESM-1b (paired $t-$ test, $p$-value $>4.2 \times 10^{-1}$). On SHS148k, KEDD achieves 6.2\% and 2.1\% absolote gains over state-of-the-art models on DFS and BFS settings (paired $t-$ test, $p$-value $<1.8 \times 10^{-2}$). It's worth noting that ESM-1b has undertaken pre-training with a vast amount of proteins, and the scale of its parameters exceeds KEDD by an order of magnitude. Thus, we expect better performance by leveraging more powerful protein sequence encoders for KEDD at the cost of extensive computation.  


 Above all, the outstanding results of KEDD indicate that structured and unstructured knowledge encapsulated within knowledge graphs and text descriptions could provide valuable biomedical insights in drug discovery. Benefiting from these knowledge, KEDD attains deep and comprehensive understanding of molecules and makes accurate predictions on a wide range of AI drug discovery tasks.

\subsection{Ablation Studies}

\subsubsection{Impact of structured and unstructured knowledge.} KEDD relies upon the integration of structured and unstructured knowledge, and we explore if these two components contributes eqally. We implement two variants of our framework, namely KEDD (w/o SK) and KEDD (w/o UK), by removing either the structured or unstructured knowledge branch. The experiment results are presented in Table \ref{tab:dp}, Table \ref{tab:ddi}, Table \ref{tab:ppi} and supplementary materials. We observe that removing either structured or unstructured knowledge leads to a significant performance drop, indicating that these two modalities are complementary with each other. Interestingly, structured knowledge plays a more significant role in relation-prediction tasks including DTI, DDI and PPI. This corroborates prior findings \cite{qiu2020gcc} that the topological information within knowledge graphs could improve the link prediction capabilities of deep learning models. On DP, unstructured knowledge brings a huge impact especially on ClinTox, indicating that molecular properties typically reside within textual descriptions. 
\begin{figure}[h]
\centering
\includegraphics[width=0.98\columnwidth]{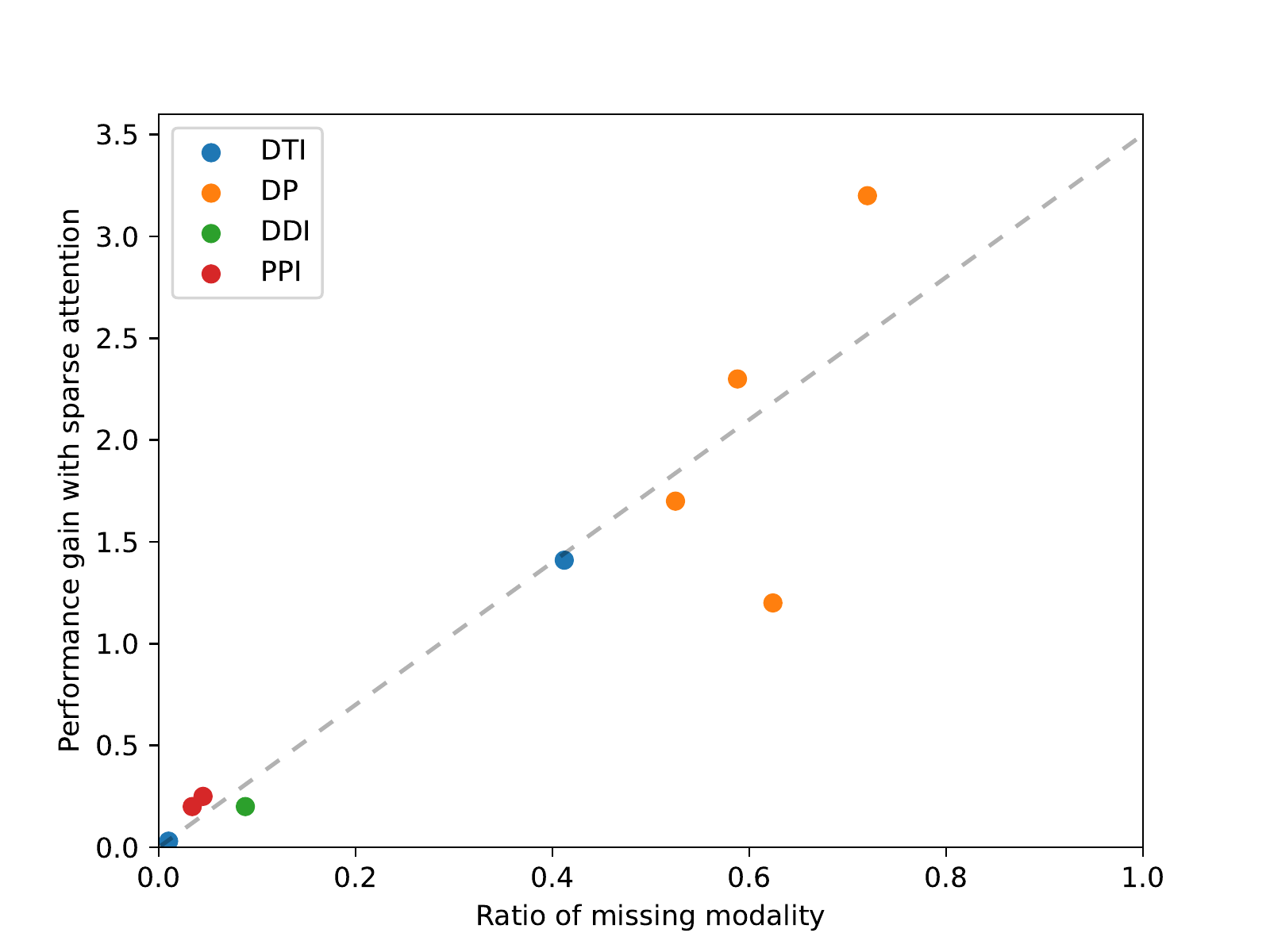}
\caption{Relationships between performance gain of sparse attention and the ratio of molecules without structured knowledge. Each dot represents the result on dataset, colored by its corresponding task.}
\label{fig:ablation}
\end{figure}

\subsubsection{Impact of sparse attention.} To investigate if the proposed sparse attention mitigates the missing modality problem, we implement KEDD (w/o SA), where we use zero vectors as $H_{X,SK}$ for drugs and proteins without structure knowledge information. We measure the severeness of missing modality by the portion of molecules without structured knowledge, and visualize its relationship with the performance gain attained by sparse attention in Figure \ref{fig:ablation}. We observe that sparse attention brings substantial improvements when encoutered with missing modalities. 


\subsubsection{Impact of modality masking.} 

KEDD proposes modality masking to obtain more training samples for sparse attention and improve robustness. We assess the impact of the masking rate  $p$ on Yamanishi08's dataset with cold drug setting. As shown in Table \ref{tab:mask}, $p = 0.05$ achieves optimal AUROC and AUPR results. When modality masking is not applied ($p=0$), the performance deteriorates by 2.4\% on average, demonstrating the significance of modality masking. However, the performance drops as $p$ continues to increase, indicating that the original structured knowledge inputs are more beneficial.

\begin{table}[h]
  \fontsize{9}{11}\selectfont
  \centering
  \begin{tabular}{ c | c c }
    \toprule
    $p$ & \textbf{AUROC} & \textbf{AUPR} \\
    \midrule
    $0.00$ & $78.0_{\pm 2.6}$ & $76.4_{\pm 2.6}$ \\
    $0.05$ & \textbf{80.4}$_{\pm \textbf{3.3}}$ & \textbf{78.7}$_{\pm \textbf{3.8}}$ \\
    $0.10$ & $80.2_{\pm 2.5}$ & $78.5_{\pm 2.9}$ \\
    $0.20$ & $79.1_{\pm 3.0}$ & $77.8_{\pm 3.4}$ \\
    \bottomrule
  \end{tabular}
  \caption{Effect of varying structured knowledge masking probability $p$ on DTI using Yamanishi08 dataset's cold drug setting.}
  \label{tab:mask}
\end{table}


\subsection{A Case Study on Real-World Drug Discovery}

\begin{figure*}[t]
\centering
\includegraphics[width=0.90\textwidth]{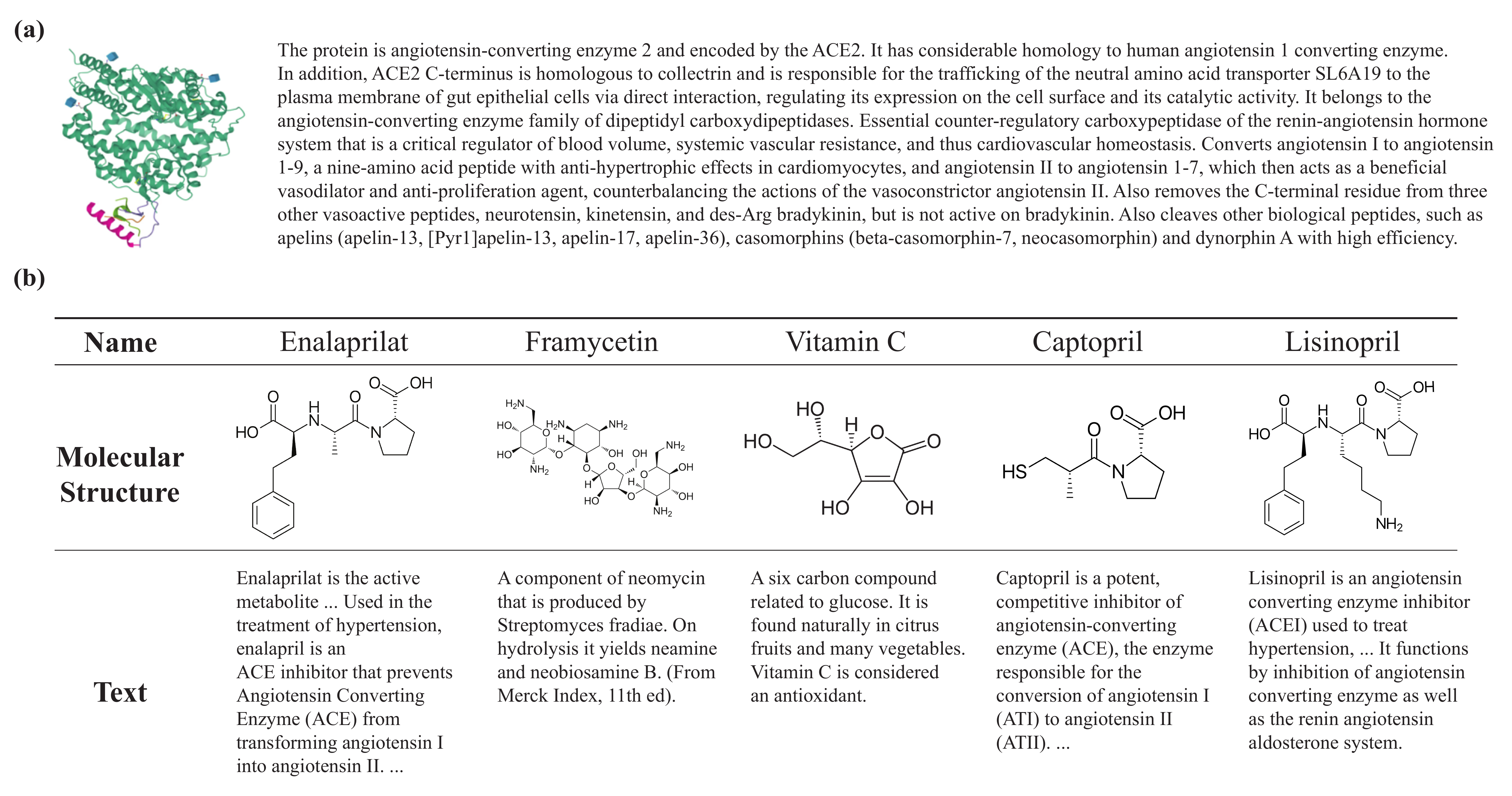} 
\caption{A drug repurposing example for ACE2. (a) Details of ACE2, a protein targeted by KEDD. (b) Top 5 drug candidates proposed by KEDD and the heterogeneous information for each.}
\label{fig:case_study}
\end{figure*}

\begin{figure}[h]
\centering
\includegraphics[width=0.95\columnwidth]{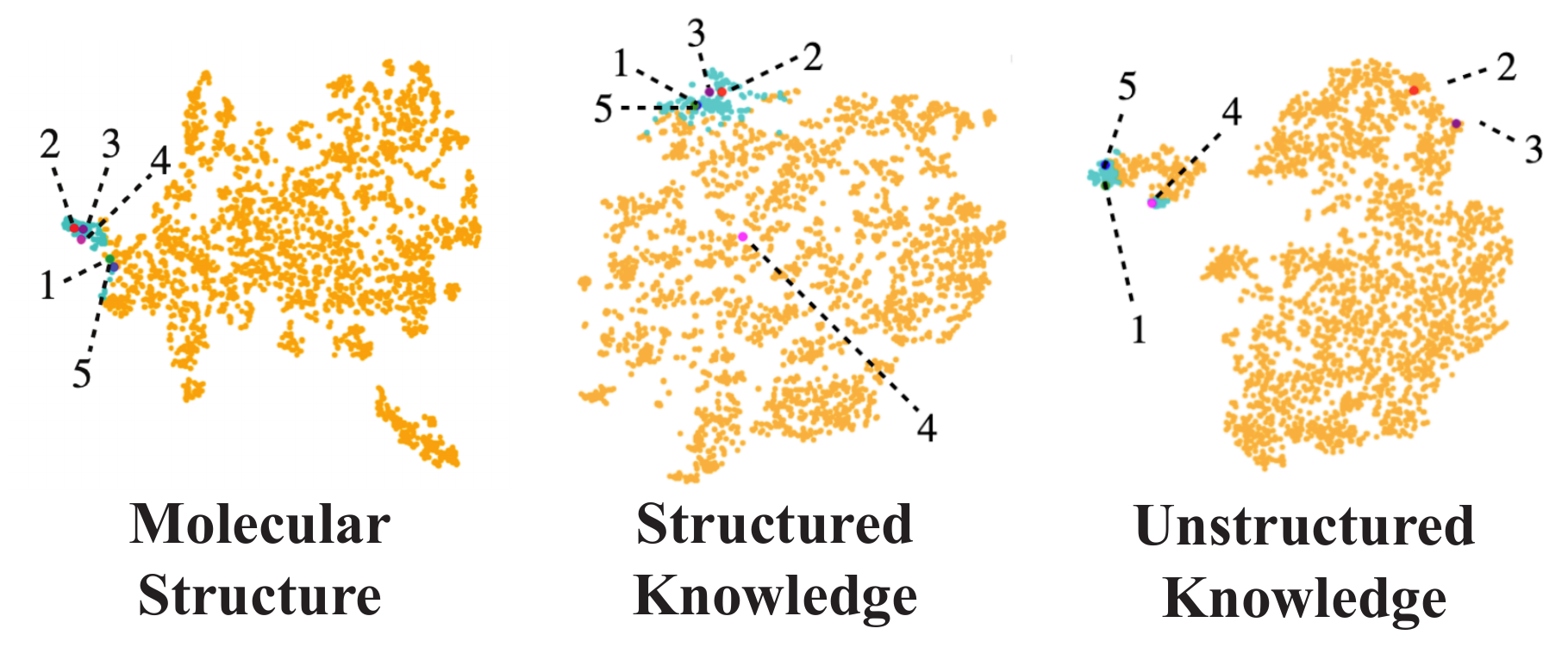} 
\caption{$t$-SNE visualization of each modality's features for drugs in BMKG. Drugs with $> 0.5$ prediction score based on each modality are highlighted, and the top-5 drug candidates for ACE2 are marked.}
\label{fig:case_study_tsne}
\end{figure}

To test the power of KEDD in real-world drug discovery scenarios, we conduct a case study on searching for drugs that bind with angiotensin-converting enzyme 2 (ACE2), a protein that has proven to be an entry receptor of SARS-CoV-2 \cite{zamorano2020ace2, li2020physiological}. We remove all data samples containing ACE2 from the BMKG-DTI dataset and train KEDD. Then, we predict the probability for each drug to interact with ACE2 and select the top 5 candidates. The heterogeneous inputs of ACE2 and each drug selected by KEDD are presented in Figure \ref{fig:case_study}. To explore the features of each modality, we visualize molecular structure, structured knowledge, and unstructured knowledge embeddings for each drug via $t$-SNE in Figure \ref{fig:case_study_tsne}.

Among the 5 drugs KEDD identified, Captopril and Lisinopril are validated active compounds, and their binding affinity values tested by wet lab experiments are reported on PubChem \cite{kim2016pubchem}. Recent studies from the biomedical domain point out that Vitamin C and Enalaprilat may have a lowering effect on the protein\cite{ivanov2021inhibition, zuo2022vitamin, moraes2021enalapril}, and an in silico work suggests that Framycetin could be a potential ACE2 inhibitor\cite{rampogu2021pharmacophore}.

As shown in Figure \ref{fig:case_study_tsne}, the molecular structure and structured knowledge features for the 5 drugs are mapped closely to each other, indicating these modalities likely played major roles in discovering the drugs. Over 99\% of the neighboring nodes of Enalaprilat and Lisinopril are the same, and their structured knowledge features are almost identical.

This case study shows that KEDD is capable of searching potential drugs for ``new targets" by comprehensively integrating structured and unstructured knowledge. Therefore, there is possibility for the framework to assist real-world drug discovery applications.

\subsection{Discussions}

While KEDD bears promise in accelerating AI drug discovery research, several efforts could be made to further extend the our framework's benefits. Firstly, the application scope of KEDD could be further extended. 3D geometries of small molecules and proteins could be incorporated as distinct modalities for biomedical insights. Other components including diseases, genes and cellular transcriptomics can also be considered. Secondly, interpretable tools that reveal the interactions between structures and sub-structures of molecules, structured knowledge and unstructured knowledge are expected in order to better assist real-world drug discovery. 



\section{Conclusion}

In this work, we present KEDD, a unified, end-to-end deep learning framework for AI drug discovery. KEDD build a novel multimodal fusion network to jointly harvest the advantages of molecular structure, structured knowledge within knowledge graphs, and unstructured knowledge within biomedical documents. To mitigate the missing modality problem of structured knowledge, KEDD leverages sparse attention as well as a modality masking technique to exploit relavant information from knowledge graphs. The effectiveness of KEDD is validated by its state-of-the-art performance on a wide spectrum of downstream tasks, including drug-target interaction prediction, drug property prediction, drug-drug interaction prediction, and protein-protein interaction. With qualitative analysis, we show KEDD's potential in assisting real-world drug discovery applications.



\bibliography{aaai24}

\end{document}